\documentclass[amsmath, amssymb, aps, prx, reprint, longbibliography, floatfix, nofootinbib]{revtex4-2}
\usepackage[utf8]{inputenc}
\usepackage[urlcolor=blue, colorlinks=true, citecolor=blue]{hyperref}

\usepackage{amsmath}
\usepackage{amsmath,amssymb}
\usepackage{nicematrix}
\usepackage{tabularx}
\usepackage{algorithm}
\usepackage{physics}
\usepackage{algpseudocode}
\usepackage{multirow}
\usepackage{adjustbox}

\usepackage{cmap}  

\begin{document}
\title{{\tt Tetra-AML}: Automatic Machine Learning via Tensor Networks}

\author{A.~Naumov}
\address{Terra Quantum AG, Kornhausstrasse 25, 9000 St. Gallen, Switzerland}

\author{Ar.~Melnikov}
\address{Terra Quantum AG, Kornhausstrasse 25, 9000 St. Gallen, Switzerland}

\author{V.~Abronin}
\address{Terra Quantum AG, Kornhausstrasse 25, 9000 St. Gallen, Switzerland}

\author{F.~Oxanichenko}
\address{Terra Quantum AG, Kornhausstrasse 25, 9000 St. Gallen, Switzerland}

\author{K.~Izmailov}
\address{Terra Quantum AG, Kornhausstrasse 25, 9000 St. Gallen, Switzerland}

\author{M.~Pflitsch}
\address{Terra Quantum AG, Kornhausstrasse 25, 9000 St. Gallen, Switzerland}

\author{A.~Melnikov}
\address{Terra Quantum AG, Kornhausstrasse 25, 9000 St. Gallen, Switzerland}

\author{M.~Perelshtein}
\email{mpe@terraquantum.swiss}
\address{Terra Quantum AG, Kornhausstrasse 25, 9000 St. Gallen, Switzerland}

\begin{abstract}

Neural networks have revolutionized many aspects of society but in the era of huge models with billions of parameters, optimizing and deploying them for commercial applications can require significant computational and financial resources.
To address these challenges, we introduce the {\tt Tetra-AML} toolbox, which automates neural architecture search and hyperparameter optimization via a custom-developed black-box {\tt Te}nsor {\tt tra}in {\tt Opt}imization algorithm, {\tt TetraOpt}.
The toolbox also provides model compression through quantization and pruning, augmented by compression using tensor networks.
Here, we analyze a unified benchmark for optimizing neural networks in computer vision tasks and show the superior performance of our approach compared to Bayesian optimization on the CIFAR-10 dataset.
We also demonstrate the compression of ResNet-18 neural networks, where we use 14.5 times less memory while losing just 3.2\% of accuracy.
The presented framework is generic, not limited by computer vision problems, supports hardware acceleration (such as with GPUs and TPUs) and can be further extended to quantum hardware and to hybrid quantum machine learning models.

\end{abstract}

\maketitle

\section*{Introduction}

Over the past decade, neural networks have influenced practically every aspect of human society \cite{ML_overview}. 
When the cutting-edge neural network AlexNet triumphed in the prestigious ImageNet challenge in the 2010s \cite{AlexNet}, the startling journey began. 
Today, AI systems like DALLE-2 that can produce realistic visuals and art from a description in natural language are still working toward the same overall goal \cite{dalle}. 
While neural networks have grown in value for both people and enterprises, putting them into practice in the context of commercial operations is getting more and more difficult every year.

The process of deploying a model involves several steps, such as model training, architecture and hyperparameter optimization, and testing, which can be computationally intensive and require significant resources. 
Furthermore, deploying a large-size model can require significant storage and computation resources, which can increase costs \cite{increasing_costs}. 
Additionally, if the model needs to be implemented on a small device with limited memory \cite{IoT}, the model may need to be optimized for size, which can add additional complexity and time to the deployment process. 
At the same time, the model's high accuracy must be maintained.
\begin{figure*}[ht]
    \centering
    \includegraphics[width = 1 \linewidth]{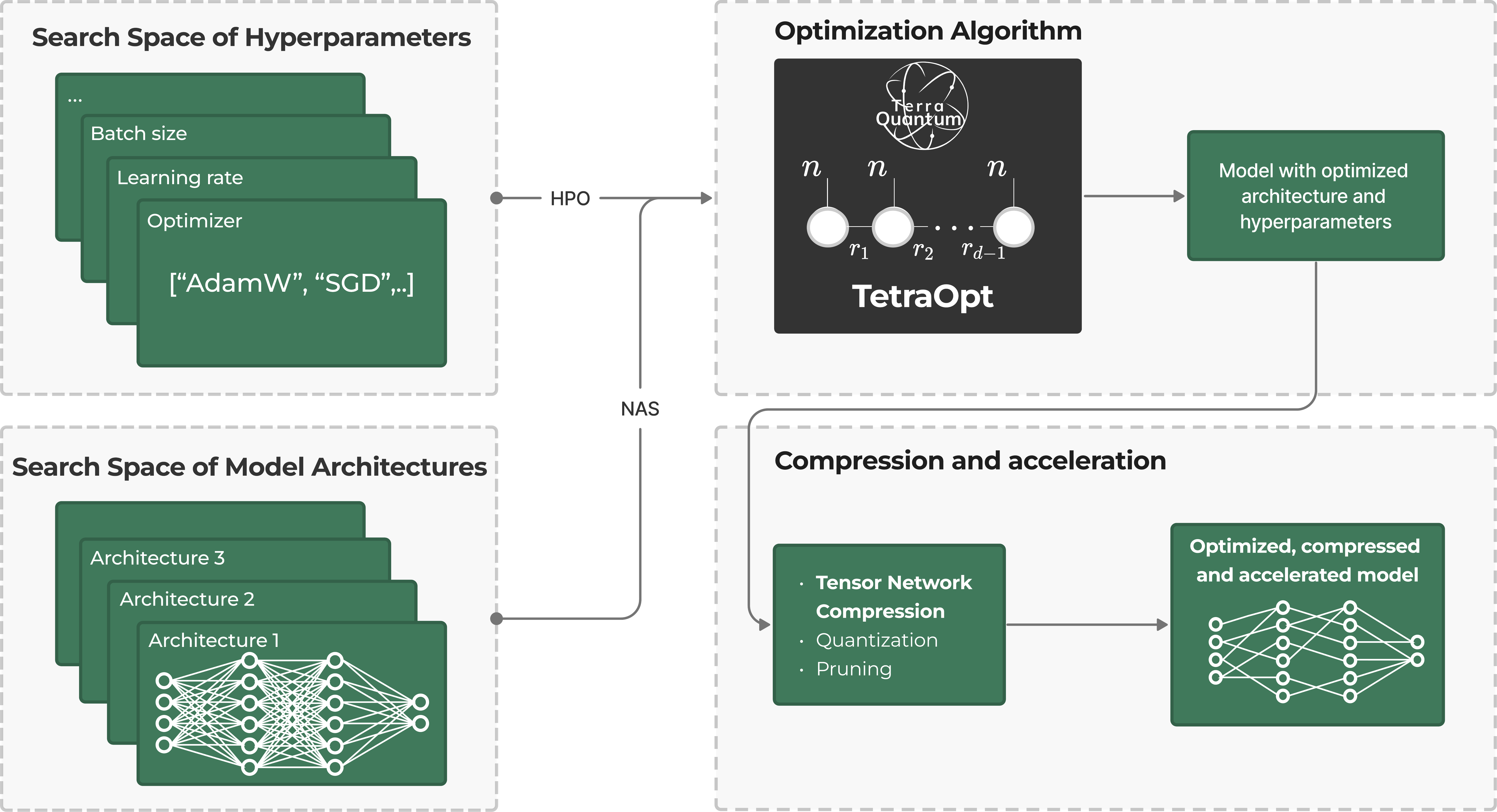}
    \caption{General scheme of {\tt Tetra-AML}. 
    Both Neural Architecture Search (NAS) and Hyperparameter Optimization (HPO) are performed via {\tt TetraOpt} (Terra Quantum's black-box optimizer based on Tensor Trains). 
    Then the model is compressed via tensor network methods, quantization and pruning.}
    \label{fig:TQ AutoML}
\end{figure*}

In addition, the practice of creating models with greater complexity in order to increase accuracy is still developing. 
There are about 60 million trainable parameters in AlexNet (2012); GPT-2 contains 1.5 billion parameters (2019); DALLE-2 and ChatGPT contain roughly 3.5 billion and 175 billion parameters, respectively, (2022), with the promise to increase the size by at least two orders of magnitude in the coming years.
It is clear that the powerful and complicated models are getting more and more expensive in terms of optimization and deployment \cite{sevilla2022compute, patterson2021carbon}.

To address these challenges, we develop an Automatic Machine Learning toolbox based on tensor networks, {\tt Tetra-AML}. 
A tensor network is a powerful numerical tool that can advance the solution of the high-dimensional problems \cite{oseledets2011tensor}.
Here we develop the framework that allows us to apply tensor networks for automatic machine learning \cite{he2021automl}, including an automatic search for the best architectures of models -- neural architecture search -- with optimal parameters -- hyperparameters optimization -- with the help of our black-box optimization approach, {\tt TetraOpt} \cite{morozov2023protein}. 
Besides, it allows compression of the models via a combination of common quantization techniques and pruning approach \cite{Quantization, pruning} augmented by compression using tensor networks \cite{novikov2015tensorizing, lee2021qttnet}. 

The general scheme of {\tt Tetra-AML} is shown in Fig.\,\ref{fig:TQ AutoML}.
{\tt Tetra-AML} offers the flexibility of bringing your own model or defining a use case to receive a suitable model. 
 A user provides a dataset and specifies the search space for optimization. 
 After that, the tool initiates parallel training of the models and applies post-training tensor network compression, pruning and quantization to create an optimal, compressed and accelerated model. 
 Once the model is ready, users can download it for deployment.
 
In this work, we mainly focus on computer vision problems, as one of the most challenging ones, but the framework is generic and can be applied for any machine learning problem. 
We consider well-known CIFAR dataset and NATS benchmarking for neural architecture search \cite{NATS}.
Besides, {\tt Tetra-AML} allows for hardware acceleration, such as with GPUs (Graphics Processing Units) and TPUs (Tensor Processing Units), and, moreover, can leverage the power of {\it quantum computers} for better optimization of classical networks and boosting the performance of quantum and hybrid quantum/classical machine learning models, which we discuss in this work.\\[35pt]


\section*{Neural Architecture Search and Hyperparameters Optimization}

The first step in building a model is choosing the best model architecture for the task at hand. 
Neural Architecture Search (NAS) is an algorithm, or set of algorithms, that automates the process of finding the best architecture for a particular ML problem and dataset \cite{NAS}. 
Such an approach explores the space of potential architectures and assesses their performance during an optimization method. 
Finding an architecture with the best performance while being computationally efficient is the ultimate aim of NAS. 
When compared to manually designing and fine-tuning architectures, this can {\bf significantly save the development time and costs} and produce new, better-performing architectures.

In addition to NAS, it is crucial to conduct hyperparameters optimization (HPO) for the model to maximize its accuracy within the specific problem and dataset. 
Hyperparameters can be considered values that are set for the model and do not change during the training regime, and may include variables such as learning rate, decay rates, choice of optimizer for the model, etc.
This tuning can be also done in an automatic manner, e.g., using black-box optimization methods \cite{falkner2018bohb}.
%
\begin{figure}[ht]
    \centering
    \includegraphics[width = 1 \linewidth]{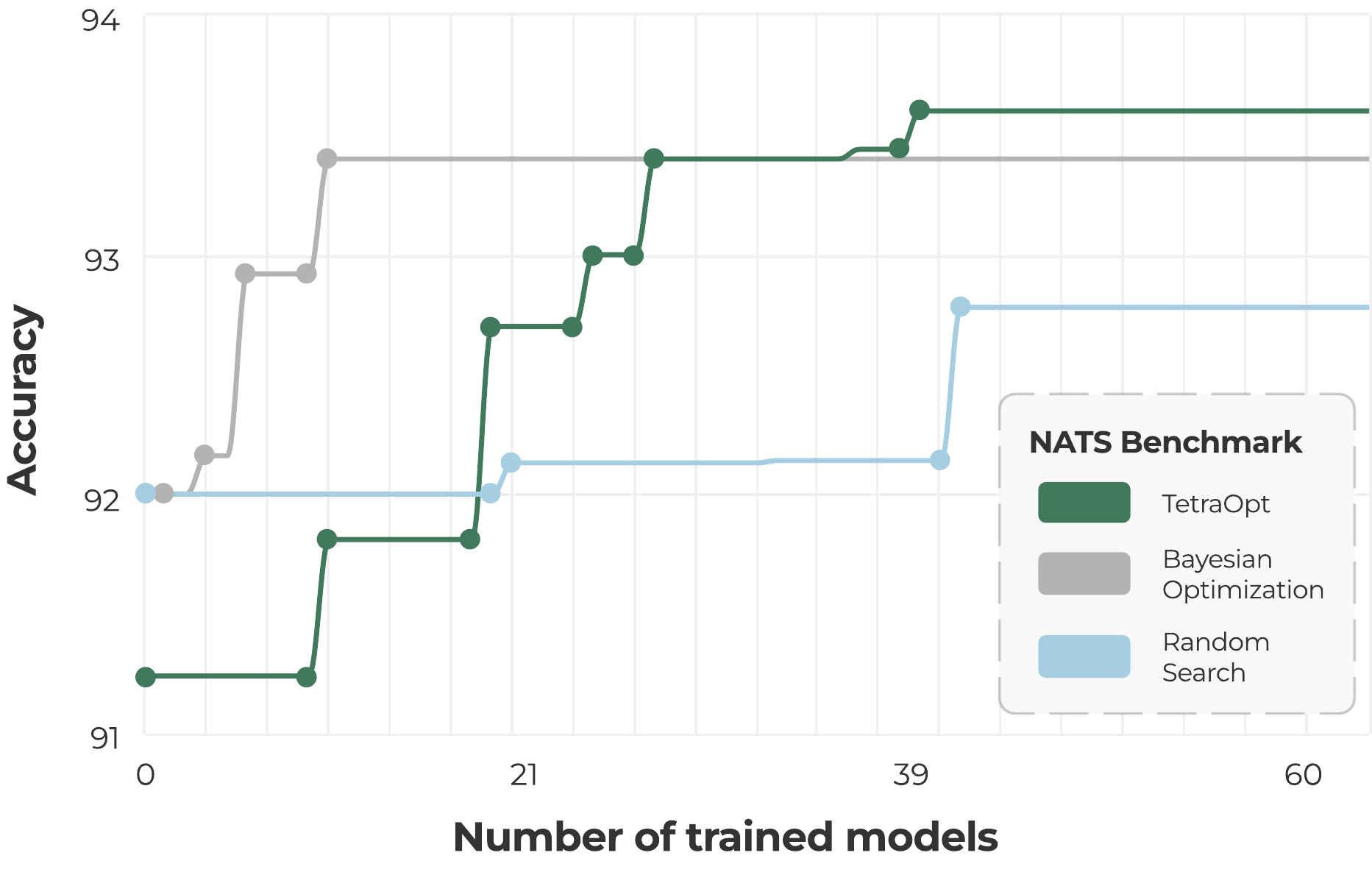}
    \caption{Validation accuracy dependence on the number of neural network models (architectures) runs for {\tt TetraOpt}, Bayesian, and Random Search algorithms. 
    {\tt TetraOpt} achieves 93.7\% accuracy, while Bayesian optimization and Random Search find architectures only with 93.5\% and 92.8\% accuracy, respectively, with the same number of model runs.
    The experiments were carried out on a standard NAS benchmark \cite{NATS}, where the CIFAR-10 dataset is used.}
    \label{fig:Nas_fig}
\end{figure}
\begin{figure*}[ht]
    \centering
    \includegraphics[width = 1 \linewidth]{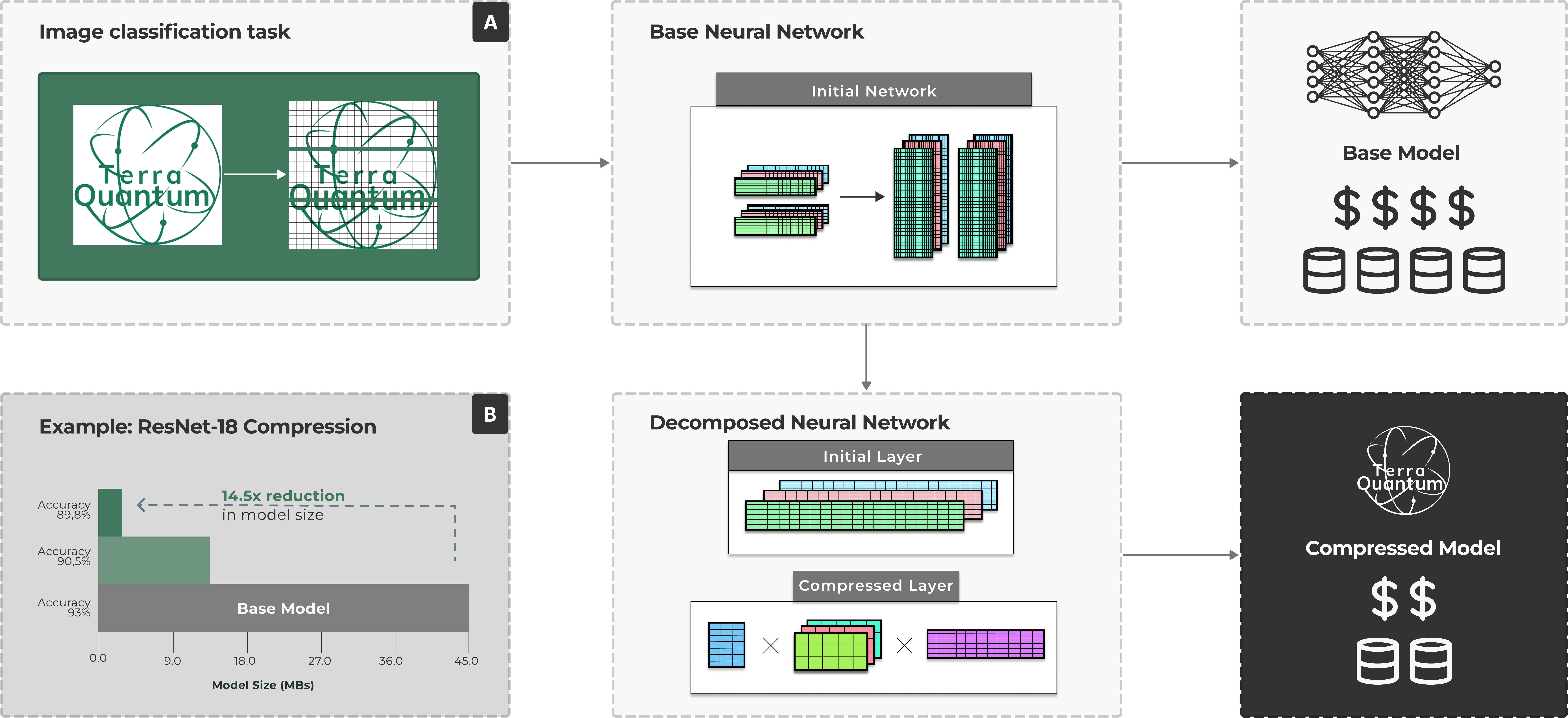}
    \caption{(A) General scheme of Neural Network compression. 
    (top) General Neural network scheme for image recognition. 
    (bottom) Compressed Convolution layer - 4D convolution kernel is represented as a sum of tensor product of small tensors (Canonical Decomposition).
    Initial layer has $C_{in} \times C_{out} \times D^2$ parameters, while after compression remains only $C_{in} \times R + D^2 \times R^2 + R \times C_{out}$. 
    For small $R$, it provides significant compression in the occupied memory.
    (B) Compression of state-of-the-art ResNet-18 on the CIFAR-10 dataset via tensor networks. 
    The diagram shows the achieved accuracy depending on the compression of the model. 
    Bottom bar: Uncompressed Base Model.
    Middle bar: TN Compressed Model (Compression coefficient: 4.5).
    Top bar: TN Compressed Model (Compression coefficient: 14.5).}
    \label{fig:compression}
\end{figure*}

Unlike other approaches, such as Random Search, Bayesian optimization methods, Reinforcement Learning or Genetic algorithms, our method utilizes a black-box optimization approach based on tensor networks, \texttt{{\tt TetraOpt}}.
{\tt TetraOpt} is a versatile global optimization algorithm that can handle various types of variables, including integers and categorical data, and is well-suited for parallel computing and hardware-acceleration via, e.g., GPUs or TPUs.
Such an optimization method is also extended to quantum processing units and can advance classical and hybrid quantum-classical machine learning models; see more discussion below.
%
%
A deep understanding of tensor networks and the ability to develop a custom optimizer advances our solution with a unique advantage in the automatic machine learning market and enables us to deliver faster and more accurate results for innovative business that actively utilizes machine learning models.

In Fig.\,\ref{fig:Nas_fig}, we present the results of a comparison between {\tt TetraOpt}, Bayesian optimization and Random Search on a standard NAS benchmark, NATS \cite{NATS}.
NATS is a well-known benchmark used to evaluate neural architecture search algorithms --  it uses the CIFAR-10 dataset.
The search space of the NATS benchmark contains a wide variety of neural network architectures, including different numbers of layers, types of layers, and connections between layers.  
It includes a search space of 15,625 neural 
architectures, which makes NATS a comprehensive and universal benchmark that enables fair comparisons between different NAS algorithms.
We observe that {\tt TetraOpt} achieves the best validation accuracy for the same number of model runs.\\[30pt]

\section*{Compression of Neural Networks using Tensor Networks}


Besides complicated tuning, the significant issue with the large models lies in the huge size of a model, which creates additional computational and financial costs that limits the efficient deployment and support.
It is even more relevant for devices with limited memory, such as local machines, mobile devices, and Internet of Things in general \cite{IoT}.
In this case, the model compression is a key technology that allows for keeping the accuracy at a decent level while substantially reducing the costs. 
While many existing approaches mostly use pruning and quantization \cite{pruning_quantization}, we focus on the development of new compression algorithms based on tensor networks \cite{novikov2015tensorizing, lebedev2014_tt_cnn}.

Tensor network compression is a cutting-edge technology that potentially offers significant advantages for businesses by compressing deep neural networks while maintaining their accuracy.
This approach reduces the number of model parameters and, consequently, the size of the model.
The main idea of this method is to decompose a huge matrix (tensor) used in a neural network layer into smaller tensors with an exponential reduction in the  memory and a substantial reduction in the runtime \cite{oseledets2011tensor}.

The crucial fact is that this technique can be applied to compress any machine learning model, including feed-forward layers \cite{novikov2015tensorizing}, recurrent neural networks \cite{tjandra2017rnn}, convolutional neural networks \cite{lebedev2014_tt_cnn}, and potentially, state-of-art transformers \cite{pham2022_transformer}. 
Moreover, by combining the powerful techniques of pruning, quantization and tensor network compression, we gain even more efficiency in resource savings \cite{lee2021qttnet}.

Here, to illustrate the capabilities of our framework we focus on the image classification problem and ResNet neural networks, which we previously used in Ref.\,\onlinecite{sagingalieva2022hyperparameter}.
We show the results of the compression of ResNet-18 on the CIFAR-10 dataset in Fig.\,\ref{fig:compression}.
During the experiment, we compress only CNN layers since they occupy an overwhelming amount of memory in the ResNet-18 architecture. 
To measure the occupied memory, we estimated the total number of parameters in the compressed and uncompressed models, taking into account fully-connected layers. 
As one can see, with a higher compression ratio, the accuracy slightly worsens, so the desired accuracy and compression might be defined by a user according to a specific problem.

\section*{Pathway to Quantum Computing}

We develop the {\tt Tetra-AML} framework keeping in mind the quantum performance enhancement that we can obtain from actively developing quantum hardware.
For instance, {\tt TetraOpt} is well extended to a quantum version, which is theoretically capable of providing more optimal points for an optimized objective -- a new optimal set of hyperparameters and neural architectures.

On the other hand, when building hybrid quantum-classical models that combine parametrized quantum circuits with classical neural networks, the process of searching for hyperparameter configurations that result in improved model accuracy and training presents a significant challenge. 
Hyperparameter Optimization and Neural Architecture Search techniques can be applied to searching the optimal parameters in Quantum Machine Learning \cite{sagingalieva2022hyperparameter} and optimal balance between classical and quantum contribution in a hybrid model. 
Such methods can be used for pure quantum models and help to find the best quantum ansatz for a given dataset or a problem. 

Overall, NAS provides a promising avenue for developing optimal model architectures for quantum and hybrid quantum computing \cite{TQ_whitepaper}, which could lead to significant advancements in the quantum machine learning field \cite{biamonte2017_QML, sagingalieva2022_QML_drug, rainjonneau2023quantum, melnikov2023_QML}.




\section*{Conclusion}

Deep Neural Networks are becoming increasingly prevalent in business as they are able to handle large and complex data sets, leading to improved performance in various tasks, such as image recognition, natural language processing and prediction. 
However, as networks become larger and more complex, the costs for development and deployment also increase.
Here, we propose an off-the-shelf solution that addresses these challenges by using tensor network techniques to find the optimal neural network architecture and hyperparameter optimization for a particular task and dataset and then compress the model to reduce its size while maintaining desired accuracy.
We develop a tool for optimizing neural networks that will significantly enhance computational efficiency, reduce the number of parameters, and ultimately improve the overall performance of deep learning models including hybrid quantum neural networks.

\bibliographystyle{ieeetr}
\bibliography{citations} 
\end{document}